\documentclass[conference]{IEEEtran}
\IEEEoverridecommandlockouts

\usepackage{cite}
\usepackage{amsmath,amssymb,amsfonts}
\usepackage{graphicx}
\usepackage{booktabs}
\usepackage{array}
\usepackage{textcomp}
\usepackage{xcolor}
\makeatletter
\g@addto@macro\normalsize{%
  \setlength\abovedisplayskip{3pt plus 1pt minus 1pt}%
  \setlength\belowdisplayskip{3pt plus 1pt minus 1pt}%
  \setlength\abovedisplayshortskip{2pt plus 1pt minus 1pt}%
  \setlength\belowdisplayshortskip{2pt plus 1pt minus 1pt}%
}
\makeatother

\begin{document}

\title{Patient-Level, Leakage-Aware Deep Learning for Cross-Center Periapical Radiograph Classification}

%---------------------------------------------------------------------
% REPLACE the author block below with the real author list.
%---------------------------------------------------------------------
\author{
\IEEEauthorblockN{Md Jubaer Rahman\textsuperscript{1,2}, Ulas Bagci\textsuperscript{2}}
\IEEEauthorblockA{
\textsuperscript{1}Department of Information and Communication Technology, Islamic University, Kushtia, Bangladesh\\
\textsuperscript{2}Hybrid and Machine Intelligence Lab, Northwestern University, Chicago, IL, USA\\
jubaerr27@gmail.com, ulas.bagci@northwestern.edu}
}

\maketitle

%=====================================================================
\begin{abstract}
Dental caries and endodontic disease are among the most common health conditions worldwide, and intraoral periapical radiographs are central to their detection, treatment planning, and follow-up. Automated tooth-level classification of these images, however, lacks reproducible benchmarks, is
often evaluated with image-level splits that leak patients between training
and test, and is rarely validated across clinics. This paper presents the
first patient-level, leakage-aware classification benchmark for single-tooth
intraoral periapical radiographs on the DentIRO dataset, which comprises
5,300 images from 3,243 patients across two clinics and four classes:
Healthy, Caries, Crowned, and Root Canal. Five transfer-learning models are
compared with patient-grouped stratified cross-validation, so that every
patient remains within a single fold. DenseNet121 gave the strongest and
most stable result at a mean macro-F1 of 0.9787, while the four
ImageNet-initialized backbones performed comparably. A controlled comparison
on a fixed architecture showed that chest-radiograph pretraining transferred
less effectively than ImageNet initialization. Bidirectional cross-center
validation produced a small average generalization gap of 0.0077, and
Grad-CAM confirmed that predictions rely on clinically meaningful tooth
regions rather than acquisition artifacts. The benchmark offers a rigorous
and reproducible baseline for intraoral radiograph classification.
\end{abstract}


%=====================================================================
\section{Introduction}

Dental caries and endodontic disease are among the most widespread health
conditions worldwide. The Global Burden of Disease Study 2021 identified
untreated caries of permanent teeth as the most prevalent of all the
conditions it tracked, affecting more than two billion people
\cite{bernabe2025gbd}. Because these conditions progress over time and are
often without symptoms in their early stages, radiographic examination
remains central to detection, treatment planning, and follow-up.

Intraoral periapical radiographs are a routine part of this examination.
They image a small number of teeth at high spatial resolution and reveal
localized detail of the crown, root, and surrounding bone that wider views
resolve less clearly. A recent systematic review and meta-analysis reported
that periapical radiography attains slightly higher diagnostic accuracy than
panoramic radiography for apical periodontitis \cite{stera2024periapical}.
This combination of resolution and tooth-level focus makes periapical images
a natural target for automated analysis, where the object of interest is a
single tooth rather than a full arch.

Deep learning has been applied to dental radiographs across a range of tasks
in recent years. Convolutional neural networks have been used for caries detection, with a 2024 meta-analysis of bitewing radiographs reporting pooled sensitivity and specificity of 0.87 and 0.89 across five studies while cautioning that most primary studies carry an unclear or high risk of bias \cite{ammar2024caries}. Systems that address several findings at once have also appeared, covering categories such as cavities, crowns, and root canal treatment that a clinician evaluates during routine review \cite{almalki2022opg}, \cite{chaudhari2025yolov8}. Much of this work is built on panoramic radiographs or framed as object detection with bounding boxes, and comparatively little targets tooth-level
classification of individual intraoral images.

Three gaps motivate the present study. First, public dental image resources
are dominated by panoramic radiographs and by annotations for segmentation or
landmark localization rather than tooth-level disease labels. A 2024 survey of
publicly available dental datasets found periapical radiographs to be a small
fraction of released data \cite{uribe2024datasets}, so reproducible
classification benchmarks on single-tooth intraoral images are scarce. Second,
many dental learning studies evaluate models with image-level random splits
even though a single patient often contributes several teeth. When images from
one patient appear in both the training and test partitions, reported
performance can be inflated beyond what the model achieves on unseen patients,
an effect measured in other medical imaging domains \cite{tampu2022leakage}.
Third, models are usually validated on data from a single source, and their
behavior across clinics with different operators and populations is rarely
reported, despite consistent evidence that accuracy declines on external data
\cite{suleman2025generalizability}.

This paper addresses these gaps using DentIRO, a recently released multi-class
dataset of single-tooth intraoral periapical radiographs labeled as Healthy,
Caries, Crowned, or Root Canal \cite{shoib2026dentiro}. We establish the first
reproducible classification benchmark on the dataset and design the evaluation
to prevent patient-level leakage. The contributions are as follows. We provide
a patient-level, leakage-aware benchmark that compares five transfer learning
models across the four diagnostic classes using stratified group
cross-validation, so that every patient stays within a single fold. We report
a controlled comparison of pretraining sources on one DenseNet121
architecture, contrasting ImageNet initialization with chest radiograph
initialization, and find that chest radiograph pretraining does not transfer
favorably to dental images. We quantify cross-center generalization through
bidirectional external validation between the two contributing centers and
report the resulting gap. Finally, we use Grad-CAM to confirm that the trained
models respond to clinically meaningful regions of the tooth rather than
background or acquisition artifacts.

%=====================================================================
\section{Related Work}

\subsection{Dental Radiograph Analysis}
Deep learning for dental radiographs has advanced quickly. On periapical
images, Chen et al. trained a region-based convolutional network to detect
caries and periapical periodontitis across several thousand radiographs
\cite{chen2021periapical}, and later work paired detector backbones with
modern convolutional designs such as ConvNeXt to localize periapical lesions
\cite{liu2024convnext}. For caries in particular, a systematic review of
network-based detection on periapical radiographs summarized steady gains in
reported accuracy alongside persistent concerns about dataset size and
evaluation protocol \cite{musri2021review}. Multi-condition systems that
resemble the label set used here have also been described. Almalki et al.
detected cavities, crowns, root canals, and broken-down roots on panoramic
images \cite{almalki2022opg}, and a YOLOv8 model classified periapical
radiographs into six categories including caries, crowns, and root canal
treatment \cite{chaudhari2025yolov8}. Related studies have modeled endodontic
treatment outcomes from two-dimensional periapical radiographs using residual
networks \cite{bennasar2025endodontic}.

Progress of this kind depends on suitable datasets, and here the situation is
more constrained. The survey noted above found that most public dental
collections rely on panoramic radiography and that periapical images form a
small share of released data \cite{uribe2024datasets}. Widely used resources
reflect this pattern. The Tufts Dental Database provides panoramic radiographs
with expert segmentation and abnormality annotations \cite{panetta2022tufts},
and panoramic benchmarks such as DENTEX supply quadrant, enumeration, and
pathosis labels for detection \cite{er2023dentex}. Recent releases continue to
favor modalities and annotation types other than tooth-level classification,
including the DenPAR periapical dataset, which supplies landmark and
segmentation labels rather than diagnostic categories
\cite{rasnayaka2025denpar}, and MMDental, a cone-beam computed tomography
collection paired with clinical records \cite{wang2025mmdental}. Single-tooth
intraoral datasets carrying disease-level labels suitable for classification
benchmarking remain uncommon, which DentIRO aims to address.

\subsection{Transfer Learning and Domain-Specific Pretraining}
Because dental datasets are modest in size, most models are initialized with
weights learned elsewhere rather than trained from scratch. Initialization
from ImageNet \cite{deng2009imagenet} is the common default and transfers
reasonably well, since early convolutional layers capture generic edge and
texture patterns. Pretraining on medical images offers an alternative. Chest
radiograph models such as CheXNet \cite{rajpurkar2017chexnet} and the
collection distributed through TorchXRayVision
\cite{cohen2021torchxrayvision} learn representations from large labeled X-ray
corpora, while RadImageNet provides a radiology-specific initialization that
has improved results on several small medical tasks relative to ImageNet
\cite{mei2022radimagenet}. The benefit is not guaranteed. Raghu et al. found
that transfer from ImageNet often yields little gain over lightweight models
trained directly on the target data, with most reuse concentrated in the
lowest layers \cite{raghu2019transfusion}, and a later analysis reported that
ImageNet initialization can match or exceed radiology-specific pretraining
across a range of tasks \cite{juodelyte2023revisiting}. Whether chest
radiograph pretraining helps for dental radiographs, whose anatomy and field
of view differ substantially from thoracic images, has not been established,
and we test this directly.

\subsection{Leakage, Validation, and Interpretability}
A dependable benchmark rests as much on evaluation design as on model choice.
When a dataset holds several images per patient, image-level random splitting
lets images from one patient fall into both training and test sets, which
inflates reported performance. The size of this effect has been measured in
other fields. Subject-level rather than slice-level splitting altered brain
MRI classification accuracy by tens of percentage points
\cite{yagis2021leakage}, and the way a longitudinal dataset is divided has
been shown to introduce identity-based shortcuts that a model can exploit
\cite{rumala2023split}. Retinal OCT experiments recorded a rise in the
Matthews correlation coefficient of up to 0.43 under improper splitting
\cite{tampu2022leakage}. DentIRO is exposed to this risk because many patients
contribute more than one tooth, which is why the present evaluation groups by
patient throughout. External validity is a second concern. Models validated at a single site frequently degrade elsewhere \cite{suleman2025generalizability}, and Youssef et al. argue that passing a single external test does not by itself establish generalizability, so validation should be repeated locally wherever a model is deployed \cite{youssef2023external}. The bidirectional evaluation reported here therefore measures transfer between two centers rather than certifying general deployability. Dental studies that validate on genuinely external data are still relatively few, with recent examples assessing caries systems on periapical and bitewing images from outside the development set \cite{szabo2024validation} and on
photographs gathered at an independent site \cite{frenkel2024external}.
Interpretability supports trust in the resulting models. Grad-CAM
\cite{selvaraju2017gradcam} highlights the regions that drive a prediction and
has been applied to dental radiographs to confirm that a network responds to
clinically relevant structures rather than incidental features
\cite{can2025explainable}. This study integrates patient-level evaluation, cross-center validation, and Grad-CAM within one benchmark.

%=====================================================================
\section{Data and Methods}

\subsection{Dataset Description}

The study uses DentIRO \cite{shoib2026dentiro}, a multi-class set of
single-tooth intraoral periapical radiographs. It comprises 5,300 grayscale
images from 3,243 patients, collected retrospectively during routine care at
two Bangladeshi clinics between 2025 and 2026: Dhaka Dental Care (HOSP1, 3,100
images) and Ma Dental Care in Cumilla (HOSP2, 2,200 images). All images were
captured with a Runyes intraoral unit and a NanoPix sensor, with exposure set
by the attending clinician, which introduces natural variation in contrast and
brightness. Informed consent, ethical approval, and de-identification were
secured before release.

Every image was screened to show a single diagnostic tooth, and overlapping teeth, motion blur, or poorly visible cases were excluded. Each image was labeled by two licensed dental practitioners under a standardized protocol during dataset creation, with disagreements resolved by consensus and CVAT used for a structural review \cite{shoib2026dentiro}. The four classes are Healthy, Caries, Crowned, and Root Canal, with 1,486, 604, 779,
and 2,431 images respectively.

Patient structure drives the evaluation design. Many patients contribute
several teeth, and 1,293 patients span more than one class, so an image-level
split would leak patients between training and test and inflate the reported
scores. The two clinics share no patients. These properties motivate the
grouped, cross-center protocols in Section III-F.

\subsection{Preprocessing and Augmentation}

All images are resized to $224 \times 224$. For the four ImageNet-initialized
backbones the grayscale image is replicated to three channels and standardized
with the ImageNet mean and standard deviation; for the chest-radiograph model
a single-channel input is scaled to the range expected by TorchXRayVision
\cite{cohen2021torchxrayvision}. Augmentation is applied only in training and
only with diagnosis-preserving transforms: rotation within
$\pm 10^{\circ}$, small affine translation and scaling, and mild brightness
and contrast jitter. Horizontal and vertical flips are excluded to preserve clinically meaningful orientations.

\subsection{Class Imbalance Handling}

Root Canal is about four times as frequent as Caries, so inverse-frequency
class weights are used, given for class $c$ by \eqref{eq:weight},

\begin{equation}
w_{c} = \frac{N}{C \cdot n_{c}}
\label{eq:weight}
\end{equation}

where $N$ is the number of training samples, $C = 4$, and $n_{c}$ is the count
of class $c$. The weights are estimated from the training partition alone and
enter the weighted cross-entropy loss in \eqref{eq:loss},

\begin{equation}
\mathcal{L} = -\frac{1}{B}\sum_{i=1}^{B} w_{y_{i}} \ln p_{i, y_{i}}
\label{eq:loss}
\end{equation}

where $B$ is the batch size, $y_{i}$ the label of sample $i$, and $p_{i,y_{i}}$ the softmax probability of that label.

\subsection{Models and Transfer Learning}

Five networks are benchmarked. Four are initialized from ImageNet
\cite{deng2009imagenet}, MobileNetV2 \cite{sandler2018mobilenetv2},
EfficientNet-B0 \cite{tan2019efficientnet}, ResNet50 \cite{he2016resnet}, and
DenseNet121 \cite{huang2017densenet}. The fifth reuses the DenseNet121
architecture but is initialized from a chest-radiograph model through
TorchXRayVision \cite{cohen2021torchxrayvision}; holding the architecture
fixed while changing the pretraining source isolates the effect of domain-specific medical pretraining. ResNet50 propagates information through residual shortcut connections and DenseNet121 through concatenation of all preceding feature maps. Each classifier head is replaced by a four-way linear layer.

Training is two-phase: the backbone is frozen while the new head is trained for three epochs at a learning rate of 0.001, then the last block is unfrozen and fine-tuned for up to fifteen epochs at 0.0001. Optimization uses AdamW
\cite{loshchilov2019adamw} with weight decay 0.0001, automatic mixed
precision, and a batch size of 32. Training stops when the validation macro-F1
does not improve for four epochs, and the best such checkpoint is retained.
All models are implemented in PyTorch.

\subsection{Evaluation Metrics}

Metrics are derived from the four-class confusion matrix. For a class $k$,
precision and recall are given in \eqref{eq:pr},

\begin{equation}
P_{k} = \frac{TP_{k}}{TP_{k} + FP_{k}}, \qquad
R_{k} = \frac{TP_{k}}{TP_{k} + FN_{k}}
\label{eq:pr}
\end{equation}

with $TP$, $FP$, and $FN$ the true positives, false positives, and false
negatives, and their harmonic mean is the class F1 score in \eqref{eq:f1},

\begin{equation}
F1_{k} = \frac{2 P_{k} R_{k}}{P_{k} + R_{k}}
\label{eq:f1}
\end{equation}

The headline metric is the macro-averaged F1 in \eqref{eq:macrof1}, which
weights every class equally and is robust to imbalance,

\begin{equation}
\text{Macro-F1} = \frac{1}{C}\sum_{k=1}^{C} F1_{k}
\label{eq:macrof1}
\end{equation}

Overall accuracy, the fraction of correctly classified images, is reported
only for completeness, whereas balanced accuracy in \eqref{eq:bacc} averages
the per-class recall,

\begin{equation}
\text{BAcc} = \frac{1}{C}\sum_{k=1}^{C} R_{k}
\label{eq:bacc}
\end{equation}

The Matthews correlation coefficient, which stays reliable under imbalance, is
computed for the multi-class case in \eqref{eq:mcc},

\begin{equation}
\text{MCC} = \frac{c \cdot s - \sum_{k} p_{k} t_{k}}
{\sqrt{\left(s^{2} - \sum_{k} p_{k}^{2}\right)
\left(s^{2} - \sum_{k} t_{k}^{2}\right)}}
\label{eq:mcc}
\end{equation}

where $c$ is the number of correct predictions, $s$ the number of samples, and
$p_{k}$ and $t_{k}$ the number of times class $k$ is predicted and truly
occurs \cite{chicco2020mcc}. Threshold-independent minority behavior is
captured by the one-vs-rest average precision in \eqref{eq:ap},

\begin{equation}
AP_{k} = \sum_{n}\left(R_{n,k} - R_{n-1,k}\right) P_{n,k}
\label{eq:ap}
\end{equation}

which is averaged over classes to give the macro PR-AUC. Every reported mean
is accompanied by a 95\% confidence interval from the Student t distribution
in \eqref{eq:ci},

\begin{equation}
\text{CI} = \bar{x} \pm t_{0.975,\,m-1} \cdot \frac{s}{\sqrt{m}}
\label{eq:ci}
\end{equation}

with $\bar{x}$ and $s$ the sample mean and standard deviation over $m$ folds
or seeds.

\subsection{Evaluation Protocols and Interpretability}

Both protocols group images by patient. Protocol A is an internal benchmark: a
five-fold stratified group cross-validation keeps
every patient inside one fold while preserving class proportions, and a
further patient-grouped 15\% of each training fold is held out for early
stopping. All five models are scored on every fold, results are reported as
the mean with the interval of \eqref{eq:ci}, and the highest mean macro-F1 in
\eqref{eq:macrof1} selects the reference model. Protocol B trains the reference model on one clinic and tests it on the whole of the other clinic in both directions, with three seeds each. Within the training clinic, a further patient-grouped 15\% split, drawn the same way as in Protocol A, is held out for early stopping. Since the clinics share no patients, the split is leakage-free. The generalization gap in \eqref{eq:gap} contrasts
the internal and external macro-F1,

\begin{equation}
\Delta = F1_{\text{int}} - F1_{\text{ext}}
\label{eq:gap}
\end{equation}

and because both clinics use identical hardware it probes robustness to
population shift rather than to device shift. Finally, Grad-CAM
\cite{selvaraju2017gradcam} applied to the reference model, using the last
dense block, verifies that predictions follow clinically meaningful regions. The full implementation will be made publicly available on GitHub upon acceptance of the manuscript.

%=====================================================================
\section{Results and Discussion}

\subsection{Internal Benchmark}

Table~\ref{tab:protocolA} reports the five-fold cross-validation results.
DenseNet121 attained the highest mean macro-F1 of 0.9787, followed by
EfficientNet-B0 at 0.9758. The four ImageNet-initialized backbones lie within
a narrow band of 0.971 to 0.979 macro-F1 and their confidence intervals
overlap, so no single natural-image backbone is decisively superior for this
task. DenseNet121 was chosen as the reference model because it also led on MCC
(0.9785), balanced accuracy (0.9764), and macro PR-AUC (0.9970). Its pooled
out-of-fold confusion matrix appears in Fig.~\ref{fig:cm}.

%---------------------------------------------------------------------
% TABLE I  (spans both columns)
%---------------------------------------------------------------------
\begin{table*}[!t]
\caption{Protocol A: Internal Patient-Level Benchmark (Mean $\pm$ 95\% CI, Five Folds)}
\label{tab:protocolA}
\centering
\begin{tabular}{lccccc}
\toprule
\textbf{Model} & \textbf{Macro-F1} & \textbf{MCC} & \textbf{Bal. Acc.} &
\textbf{PR-AUC} & \textbf{Accuracy} \\
\midrule
MobileNetV2 & 0.9715 $\pm$ 0.0130 & 0.9715 $\pm$ 0.0126 & 0.9710 $\pm$ 0.0179 & 0.9942 $\pm$ 0.0048 & 0.9806 $\pm$ 0.0088 \\
EfficientNet-B0 & 0.9758 $\pm$ 0.0073 & 0.9754 $\pm$ 0.0065 & 0.9739 $\pm$ 0.0111 & 0.9954 $\pm$ 0.0013 & 0.9834 $\pm$ 0.0044 \\
ResNet50 & 0.9716 $\pm$ 0.0034 & 0.9713 $\pm$ 0.0020 & 0.9675 $\pm$ 0.0080 & 0.9943 $\pm$ 0.0018 & 0.9806 $\pm$ 0.0016 \\
DenseNet121 & 0.9787 $\pm$ 0.0043 & 0.9785 $\pm$ 0.0045 & 0.9764 $\pm$ 0.0035 & 0.9970 $\pm$ 0.0012 & 0.9855 $\pm$ 0.0029 \\
DenseNet121-XRV & 0.9367 $\pm$ 0.0191 & 0.9349 $\pm$ 0.0205 & 0.9375 $\pm$ 0.0212 & 0.9783 $\pm$ 0.0123 & 0.9558 $\pm$ 0.0140 \\
\bottomrule
\end{tabular}
\end{table*}

\subsection{Effect of the Pretraining Source}

The two DenseNet121 rows of Table~\ref{tab:protocolA} isolate the pretraining
source, since the architecture is fixed. The chest-radiograph variant
(DenseNet121-XRV) reached only 0.9367 macro-F1, which is 0.0420 below the
ImageNet-initialized model and well outside the overlapping band of the
natural-image networks. Chest and intraoral radiographs differ substantially
in anatomy, field of view, and texture, so thoracic features transfer less
effectively than the broader ImageNet representations. For this task,
domain-specific medical pretraining did not help, consistent with reports that
ImageNet initialization can match or exceed radiology-specific pretraining
\cite{raghu2019transfusion}, \cite{juodelyte2023revisiting}.

\subsection{Per-Class Performance}

\begin{figure}[!t]
\centering
\includegraphics[width=\columnwidth]{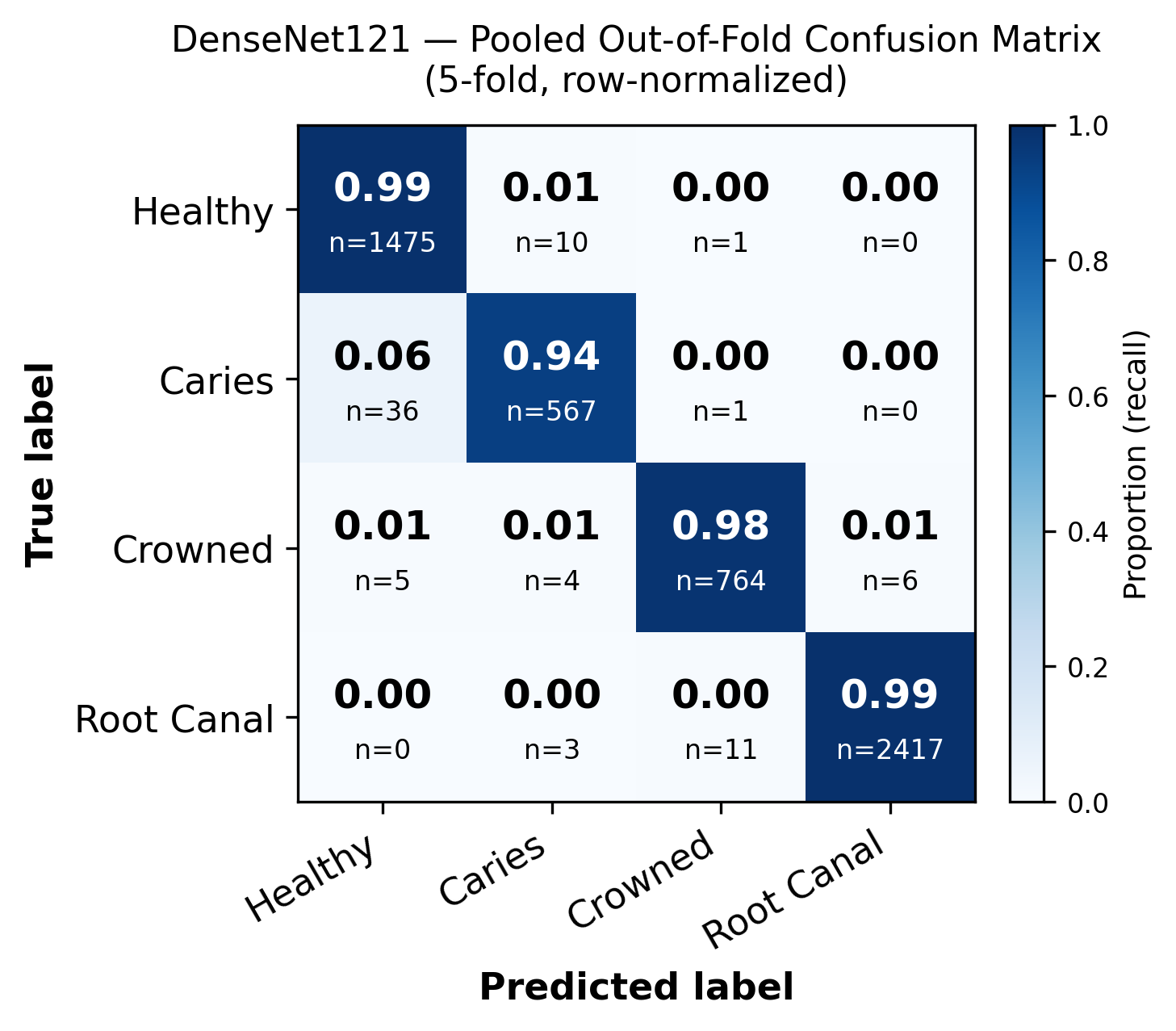}
\caption{Pooled out-of-fold confusion matrix of DenseNet121 (row-normalized).}
\label{fig:cm}
\end{figure}

Table~\ref{tab:perclass} details the per-class behavior of DenseNet121.
Healthy, Crowned, and Root Canal are recognized almost perfectly, with F1
scores of 0.9826, 0.9819, and 0.9959, and Root Canal is strongest because
endodontic filling is markedly radio-opaque. Caries is the hardest category,
with the lowest recall of 0.9383 against a high precision of 0.9710: the model
rarely labels other teeth as carious but misses a fraction of true caries,
which is the clinically more sensitive error. This is consistent with the
small size of the class and the subtle appearance of early decay relative to
the bright markers of the restorative classes. The same confusions are visible
off the diagonal of Fig.~\ref{fig:cm}.

%---------------------------------------------------------------------
% TABLE II  (spans both columns)
%---------------------------------------------------------------------
\begin{table*}[!t]
\caption{Per-Class Performance of DenseNet121 (Out-of-Fold, Mean $\pm$ 95\% CI)}
\label{tab:perclass}
\centering
\begin{tabular}{lccccc}
\toprule
\textbf{Class} & \textbf{Support} & \textbf{Precision} & \textbf{Recall} &
\textbf{F1} & \textbf{PR-AUC} \\
\midrule
Healthy & 1486 & 0.9731 $\pm$ 0.0033 & 0.9924 $\pm$ 0.0067 & 0.9826 $\pm$ 0.0024 & 0.9985 $\pm$ 0.0004 \\
Caries & 604 & 0.9710 $\pm$ 0.0073 & 0.9383 $\pm$ 0.0094 & 0.9543 $\pm$ 0.0049 & 0.9916 $\pm$ 0.0024 \\
Crowned & 779 & 0.9832 $\pm$ 0.0198 & 0.9807 $\pm$ 0.0057 & 0.9819 $\pm$ 0.0113 & 0.9981 $\pm$ 0.0023 \\
Root Canal & 2431 & 0.9975 $\pm$ 0.0012 & 0.9944 $\pm$ 0.0088 & 0.9959 $\pm$ 0.0043 & 0.9999 $\pm$ 0.0001 \\
\bottomrule
\end{tabular}
\end{table*}

\subsection{Cross-Center Generalization}

\begin{figure}[htbp]
    \centering
    \includegraphics[width=\linewidth]{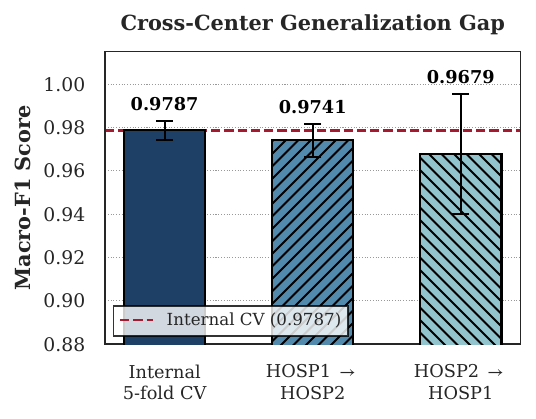}
    \caption{Bidirectional cross-center validation of DenseNet121, displaying a minimal average generalization gap when tested on external clinic data.}
    \label{fig:cross_center}
\end{figure}

Table~\ref{tab:protocolB} summarizes the bidirectional cross-center validation
of DenseNet121, plotted against the internal reference in
Fig.~\ref{fig:cross_center}. Training on Dhaka and testing on Cumilla (HOSP1 to
HOSP2) lowered macro-F1 by only 0.0046 relative to the internal
cross-validation, and the reverse direction by 0.0108, for an average
generalization gap of 0.0077. Performance therefore transfers well across the
two centers. The wider interval of the HOSP2 to HOSP1 direction
($\pm$0.0277) reflects the smaller training set of Cumilla (2,200 images),
which raises seed-to-seed variance rather than indicating a systematic bias.
Because Caries is the only class with meaningful headroom, its recall is the more informative signal: it fell only slightly to $0.9501 \pm 0.0402$ from HOSP1 to HOSP2 but dropped to $0.8838 \pm 0.1464$ from HOSP2 to HOSP1, a larger decline than the overall macro-F1 gap suggests.

%---------------------------------------------------------------------
% TABLE III  (single column)
%---------------------------------------------------------------------
\begin{table}[!t]
\caption{Protocol B: Bidirectional Cross-Center Validation of DenseNet121}
\label{tab:protocolB}
\centering
\setlength{\tabcolsep}{3pt}
\resizebox{\columnwidth}{!}{%
\begin{tabular}{lcccc}
\toprule
\textbf{Direction} & \textbf{Macro-F1} & \textbf{MCC} & \textbf{Gen. Gap} & \textbf{Caries Rec.} \\
\midrule
HOSP1 $\rightarrow$ HOSP2 & 0.9741 $\pm$ 0.0076 & 0.9746 $\pm$ 0.0071 & +0.0046 & 0.9501 $\pm$ 0.0402 \\
HOSP2 $\rightarrow$ HOSP1 & 0.9679 $\pm$ 0.0277 & 0.9683 $\pm$ 0.0249 & +0.0108 & 0.8838 $\pm$ 0.1464 \\
\bottomrule
\end{tabular}%
}
\end{table}

\subsection{Interpretability}

\begin{figure}[!t]
\centering
\includegraphics[width=\columnwidth]{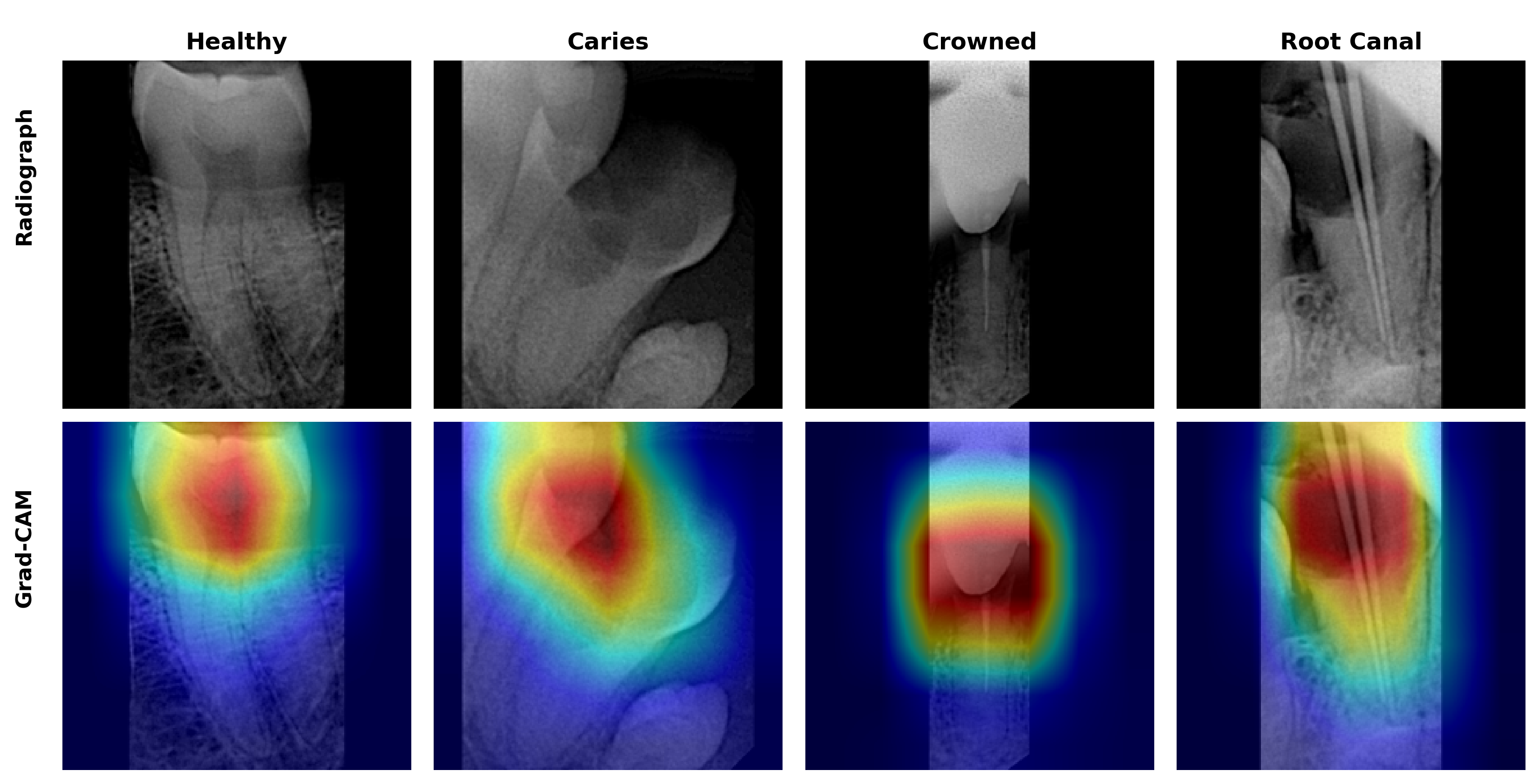}
\caption{Grad-CAM heatmaps for one representative image per class,
highlighting class-discriminative regions.}
\label{fig:gradcam}
\end{figure}

Fig.~\ref{fig:gradcam} shows Grad-CAM heatmaps for one representative image
per class. For the restorative classes the activation concentrates on the
crown and on the root-canal filling, the exact regions a clinician inspects,
whereas for Healthy and Caries the model attends to the overall tooth
morphology in the absence of a bright marker. The maps indicate that the high
accuracy arises from clinically meaningful structures rather than from
background or acquisition artifacts.

\subsection{Limitations}

Several limitations qualify these results. Both clinics used the same Runeys
and NanoPix system, so the cross-center evaluation measures robustness to
population shift but not to device-level domain shift, which is often the
larger source of degradation. The four natural-image backbones have overlapping intervals, so their ranking should be read as a tie rather than a strict order. The data come from two clinics in one country, and labels reflect a two-reader consensus rather than histopathological confirmation.
In addition, resizing high-resolution periapical images to 224×224 for the network input may discard fine detail that early caries lesions depend on, which could partly explain the lower Caries recall.
Finally, the lower recall of Caries means a fraction of early lesions is still missed, which matters most in screening.

\subsection{Future Work}

Future work will extend the benchmark to centers with different imaging
hardware to test device-level generalization directly, enlarge the Caries
class to close its recall gap, and add lesion localization and calibrated
probabilities to support clinical use. Semi-supervised and self-supervised
pretraining on unlabeled intraoral radiographs is a further direction given
the limited size of labeled dental datasets.

%=====================================================================
\section{Conclusions}

This paper introduced the first reproducible, patient-level, leakage-aware
classification benchmark for single-tooth intraoral periapical radiographs on
the DentIRO dataset. Across five transfer-learning backbones evaluated with
patient-grouped stratified cross-validation, DenseNet121 gave the strongest
and most stable performance at 0.9787 macro-F1, while the four natural-image
backbones performed comparably and chest-radiograph pretraining transferred
less effectively than ImageNet initialization. Bidirectional cross-center
validation showed a small average generalization gap of 0.0077, and Grad-CAM
confirmed that predictions rely on clinically meaningful regions. Because both
centers shared the same acquisition hardware, this cross-center result should
be read as robustness to population shift; validating across different devices
and strengthening the minority Caries class are the main steps toward clinical
deployment.

% নতুন
\section*{Acknowledgment}
The authors used Claude (Anthropic) for language editing and formatting; the experiments, analysis, and interpretation are the authors' own.

%=====================================================================
\bibliographystyle{IEEEtran}
\bibliography{references}

@article{bernabe2025gbd,
  author  = {E. Bernabe and others},
  title   = {Trends in the global, regional, and national burden of oral conditions from 1990 to 2021: a systematic analysis for the {Global Burden of Disease Study 2021}},
  journal = {The Lancet},
  volume  = {405},
  number  = {10482},
  pages   = {897--910},
  year    = {2025},
  doi     = {10.1016/S0140-6736(24)02811-3}
}

@article{stera2024periapical,
  author  = {G. Stera and M. Giusti and A. Magnini and L. Calistri and R. Izzetti and C. Nardi},
  title   = {Diagnostic accuracy of periapical radiography and panoramic radiography in the detection of apical periodontitis: a systematic review and meta-analysis},
  journal = {La Radiologia Medica},
  volume  = {129},
  number  = {11},
  pages   = {1682--1695},
  year    = {2024},
  doi     = {10.1007/s11547-024-01882-z}
}

@article{ammar2024caries,
  author  = {N. Ammar and J. K{\"u}hnisch},
  title   = {Diagnostic performance of artificial intelligence-aided caries detection on bitewing radiographs: a systematic review and meta-analysis},
  journal = {Japanese Dental Science Review},
  volume  = {60},
  pages   = {128--136},
  year    = {2024},
  doi     = {10.1016/j.jdsr.2024.02.001}
}

@article{almalki2022opg,
  author  = {Y. E. Almalki and others},
  title   = {Deep learning models for classification of dental diseases using orthopantomography {X}-ray {OPG} images},
  journal = {Sensors},
  volume  = {22},
  number  = {19},
  pages   = {7370},
  year    = {2022},
  doi     = {10.3390/s22197370}
}

@article{chaudhari2025yolov8,
  author  = {A. Chaudhari and P. Birwadkar and S. V. Joshi and Y. Verma and R. Sindgi},
  title   = {Classification of periapical dental {X}-ray using the {YOLOv8} deep learning model},
  journal = {MethodsX},
  volume  = {15},
  pages   = {103721},
  year    = {2025},
  doi     = {10.1016/j.mex.2025.103721}
}

@article{uribe2024datasets,
  author  = {S. E. Uribe and others},
  title   = {Publicly available dental image datasets for artificial intelligence},
  journal = {Journal of Dental Research},
  volume  = {103},
  number  = {13},
  pages   = {1365--1374},
  year    = {2024},
  doi     = {10.1177/00220345241272052}
}

@article{tampu2022leakage,
  author  = {I. E. Tampu and A. Eklund and N. Haj-Hosseini},
  title   = {Inflation of test accuracy due to data leakage in deep learning-based classification of {OCT} images},
  journal = {Scientific Data},
  volume  = {9},
  number  = {1},
  pages   = {580},
  year    = {2022},
  doi     = {10.1038/s41597-022-01618-6}
}

@article{suleman2025generalizability,
  author  = {M. U. Suleman and others},
  title   = {Assessing the generalizability of artificial intelligence in radiology: a systematic review of performance across different clinical settings},
  journal = {Annals of Medicine and Surgery},
  volume  = {87},
  number  = {12},
  pages   = {8803--8811},
  year    = {2025},
  doi     = {10.1097/MS9.0000000000004166}
}

@misc{shoib2026dentiro,
  author       = {Md. M. H. Shoib and others},
  title        = {{DentIRO}: A high-quality multi-class single-tooth intraoral radiograph dataset for automated dental diagnosis},
  howpublished = {Figshare},
  year         = {2026},
  doi          = {10.6084/m9.figshare.32086377.v1}
}

@article{chen2021periapical,
  author  = {H. Chen and H. Li and Y. Zhao and J. Zhao and Y. Wang},
  title   = {Dental disease detection on periapical radiographs based on deep convolutional neural networks},
  journal = {International Journal of Computer Assisted Radiology and Surgery},
  volume  = {16},
  number  = {4},
  pages   = {649--661},
  year    = {2021},
  doi     = {10.1007/s11548-021-02319-y}
}

@article{liu2024convnext,
  author  = {J. Liu and others},
  title   = {Periapical lesion detection in periapical radiographs using the latest convolutional neural network {ConvNeXt} and its integrated models},
  journal = {Scientific Reports},
  volume  = {14},
  number  = {1},
  pages   = {25429},
  year    = {2024},
  doi     = {10.1038/s41598-024-75748-9}
}

@article{musri2021review,
  author  = {N. Musri and B. Christie and S. J. A. Ichwan and A. Cahyanto},
  title   = {Deep learning convolutional neural network algorithms for the early detection and diagnosis of dental caries on periapical radiographs: a systematic review},
  journal = {Imaging Science in Dentistry},
  volume  = {51},
  number  = {3},
  pages   = {237},
  year    = {2021},
  doi     = {10.5624/isd.20210074}
}

@article{bennasar2025endodontic,
  author  = {C. Bennasar and A. Nadal-Mart{\'i}nez and S. Arroyo and Y. Gonzalez-Cid and {\'A}. A. L{\'o}pez-Gonz{\'a}lez and P. J. T{\'a}rraga},
  title   = {Integrating machine learning and deep learning for predicting non-surgical root canal treatment outcomes using two-dimensional periapical radiographs},
  journal = {Diagnostics},
  volume  = {15},
  number  = {8},
  pages   = {1009},
  year    = {2025},
  doi     = {10.3390/diagnostics15081009}
}

@article{panetta2022tufts,
  author  = {K. Panetta and R. Rajendran and A. Ramesh and S. Rao and S. Agaian},
  title   = {{Tufts Dental Database}: a multimodal panoramic {X}-ray dataset for benchmarking diagnostic systems},
  journal = {IEEE Journal of Biomedical and Health Informatics},
  volume  = {26},
  number  = {4},
  pages   = {1650--1659},
  year    = {2022},
  doi     = {10.1109/JBHI.2021.3117575}
}

@article{er2023dentex,
  author  = {I. E. Hamamci and others},
  title   = {{DENTEX}: dental enumeration and tooth pathosis detection benchmark for panoramic {X}-ray},
  journal = {arXiv preprint arXiv:2305.19112},
  year    = {2023},
  doi     = {10.48550/arXiv.2305.19112}
}

@article{rasnayaka2025denpar,
  author  = {S. Rasnayaka and others},
  title   = {{DenPAR}: annotated intra-oral periapical radiographs dataset for machine learning},
  journal = {Scientific Data},
  volume  = {12},
  number  = {1},
  pages   = {1615},
  year    = {2025},
  doi     = {10.1038/s41597-025-05906-9}
}

@article{wang2025mmdental,
  author  = {C. Wang and others},
  title   = {{MMDental}: a multimodal dataset of tooth {CBCT} images with expert medical records},
  journal = {Scientific Data},
  volume  = {12},
  number  = {1},
  pages   = {1172},
  year    = {2025},
  doi     = {10.1038/s41597-025-05398-7}
}

@inproceedings{deng2009imagenet,
  author    = {J. Deng and W. Dong and R. Socher and L.-J. Li and K. Li and L. Fei-Fei},
  title     = {{ImageNet}: a large-scale hierarchical image database},
  booktitle = {Proc. IEEE Conf. Comput. Vis. Pattern Recognit. (CVPR)},
  pages     = {248--255},
  year      = {2009},
  doi       = {10.1109/CVPR.2009.5206848}
}

@article{rajpurkar2017chexnet,
  author  = {P. Rajpurkar and others},
  title   = {{CheXNet}: radiologist-level pneumonia detection on chest {X}-rays with deep learning},
  journal = {arXiv preprint arXiv:1711.05225},
  year    = {2017},
  doi     = {10.48550/arXiv.1711.05225}
}

@article{cohen2021torchxrayvision,
  author  = {J. P. Cohen and others},
  title   = {{TorchXRayVision}: a library of chest {X}-ray datasets and models},
  journal = {arXiv preprint arXiv:2111.00595},
  year    = {2021},
  doi     = {10.48550/arXiv.2111.00595}
}

@article{mei2022radimagenet,
  author  = {X. Mei and others},
  title   = {{RadImageNet}: an open radiologic deep learning research dataset for effective transfer learning},
  journal = {Radiology: Artificial Intelligence},
  volume  = {4},
  number  = {5},
  pages   = {e210315},
  year    = {2022},
  doi     = {10.1148/ryai.210315}
}

@inproceedings{raghu2019transfusion,
  author    = {M. Raghu and C. Zhang and J. Kleinberg and S. Bengio},
  title     = {Transfusion: understanding transfer learning for medical imaging},
  booktitle = {Adv. Neural Inf. Process. Syst. (NeurIPS)},
  volume    = {32},
  pages     = {3347--3357},
  year      = {2019}
}

@article{juodelyte2023revisiting,
  author  = {D. Juodelyte and A. Jim{\'e}nez-S{\'a}nchez and V. Cheplygina},
  title   = {Revisiting hidden representations in transfer learning for medical imaging},
  journal = {Transactions on Machine Learning Research},
  year    = {2023}
}

@article{yagis2021leakage,
  author  = {E. Yagis and others},
  title   = {Effect of data leakage in brain {MRI} classification using {2D} convolutional neural networks},
  journal = {Scientific Reports},
  volume  = {11},
  number  = {1},
  pages   = {22544},
  year    = {2021},
  doi     = {10.1038/s41598-021-01681-w}
}

@incollection{rumala2023split,
  author    = {D. J. Rumala},
  title     = {How you split matters: data leakage and subject characteristics studies in longitudinal brain {MRI} analysis},
  booktitle = {Lecture Notes in Computer Science},
  pages     = {235--245},
  year      = {2023},
  publisher = {Springer},
  doi       = {10.1007/978-3-031-45249-9\_23}
}

@article{youssef2023external,
  author  = {A. Youssef and M. Pencina and A. Thakur and T. Zhu and D. Clifton and N. H. Shah},
  title   = {External validation of {AI} models in health should be replaced with recurring local validation},
  journal = {Nature Medicine},
  volume  = {29},
  number  = {11},
  pages   = {2686--2687},
  year    = {2023},
  doi     = {10.1038/s41591-023-02540-z}
}

@article{szabo2024validation,
  author  = {V. Szab{\'o} and others},
  title   = {Validation of artificial intelligence application for dental caries diagnosis on intraoral bitewing and periapical radiographs},
  journal = {Journal of Dentistry},
  volume  = {147},
  pages   = {105105},
  year    = {2024},
  doi     = {10.1016/j.jdent.2024.105105}
}

@article{frenkel2024external,
  author  = {E. Frenkel and others},
  title   = {Caries detection and classification in photographs using an artificial intelligence-based model: an external validation study},
  journal = {Diagnostics},
  volume  = {14},
  number  = {20},
  pages   = {2281},
  year    = {2024},
  doi     = {10.3390/diagnostics14202281}
}

@inproceedings{selvaraju2017gradcam,
  author    = {R. R. Selvaraju and M. Cogswell and A. Das and R. Vedantam and D. Parikh and D. Batra},
  title     = {{Grad-CAM}: visual explanations from deep networks via gradient-based localization},
  booktitle = {Proc. IEEE Int. Conf. Comput. Vis. (ICCV)},
  pages     = {618--626},
  year      = {2017},
  doi       = {10.1109/ICCV.2017.74}
}

@article{can2025explainable,
  author  = {Z. Can and E. C. Ayd{\i}n},
  title   = {Explainable {CNN}-radiomics fusion and ensemble learning for multimodal lesion classification in dental radiographs},
  journal = {Diagnostics},
  volume  = {15},
  number  = {16},
  pages   = {1997},
  year    = {2025},
  doi     = {10.3390/diagnostics15161997}
}

@inproceedings{sandler2018mobilenetv2,
  author    = {M. Sandler and A. Howard and M. Zhu and A. Zhmoginov and L.-C. Chen},
  title     = {{MobileNetV2}: inverted residuals and linear bottlenecks},
  booktitle = {Proc. IEEE Conf. Comput. Vis. Pattern Recognit. (CVPR)},
  pages     = {4510--4520},
  year      = {2018},
  doi       = {10.1109/CVPR.2018.00474}
}

@inproceedings{tan2019efficientnet,
  author    = {Tan, Mingxing and Le, Quoc V.},
  title     = {{EfficientNet}: Rethinking Model Scaling for Convolutional Neural Networks},
  booktitle = {Proceedings of the 36th International Conference on Machine Learning},
  series    = {Proceedings of Machine Learning Research},
  volume    = {97},
  pages     = {6105--6114},
  year      = {2019},
  publisher = {PMLR},
}

@inproceedings{he2016resnet,
  author    = {K. He and X. Zhang and S. Ren and J. Sun},
  title     = {Deep residual learning for image recognition},
  booktitle = {Proc. IEEE Conf. Comput. Vis. Pattern Recognit. (CVPR)},
  pages     = {770--778},
  year      = {2016}
}

@inproceedings{huang2017densenet,
  author    = {G. Huang and Z. Liu and L. van der Maaten and K. Q. Weinberger},
  title     = {Densely connected convolutional networks},
  booktitle = {Proc. IEEE Conf. Comput. Vis. Pattern Recognit. (CVPR)},
  pages     = {4700--4708},
  year      = {2017}
}

@inproceedings{loshchilov2019adamw,
  author    = {I. Loshchilov and F. Hutter},
  title     = {Decoupled weight decay regularization},
  booktitle = {Proc. Int. Conf. Learn. Represent. (ICLR)},
  year      = {2019}
}

@article{chicco2020mcc,
  author  = {D. Chicco and G. Jurman},
  title   = {The advantages of the {Matthews} correlation coefficient ({MCC}) over {F1} score and accuracy in binary classification evaluation},
  journal = {BMC Genomics},
  volume  = {21},
  pages   = {6},
  year    = {2020},
  doi     = {10.1186/s12864-019-6413-7}
}

\end{document}